\documentclass[11pt,letterpaper]{article}

\usepackage[title]{appendix}

\usepackage{caption}
\usepackage{subcaption}
\usepackage{fullpage}
\usepackage{times}
\usepackage{array}
\usepackage{ragged2e}
\usepackage{multirow}
\usepackage{fancyhdr, amssymb, mathtools}
\usepackage[ruled,vlined]{algorithm2e}
\usepackage[dvipsnames,table]{xcolor}
\usepackage[margin = 1 in]{geometry}
\usepackage{graphicx}
\usepackage{outlines}
\usepackage{amsmath}

\DeclareMathOperator*{\argmin}{argmin}
\usepackage{graphicx}
\usepackage{dirtytalk}

\usepackage{multirow}
\usepackage{arydshln}

\usepackage[english]{babel} 
\usepackage{blindtext}
\usepackage{amsfonts}
\usepackage{amssymb}
\usepackage{bbm}
\usepackage{mathtools}

\usepackage{hhline}
\usepackage{makecell}

\usepackage{xltabular}
\usepackage{arydshln} 
\usepackage{wrapfig}
\usepackage{enumitem}
\usepackage{floatrow}
\usepackage{xfrac}
\usepackage[font={footnotesize}, labelfont={bf}, labelsep=space, justification=justified, skip=0pt]{caption}
\usepackage[hang,flushmargin]{footmisc} 
\usepackage[colorlinks=true, allcolors=blue]{hyperref}
\usepackage{booktabs}
\usepackage{csquotes}

\definecolor{headcolor}{RGB}{255, 255, 255}
\definecolor{columncolor}{RGB}{255, 255, 255}
\definecolor{modelcolor}{RGB}{255, 255, 255}
\definecolor{optioncolor}{RGB}{255, 255, 255}

\definecolor{codegreen}{rgb}{0,0.6,0}
\definecolor{codegray}{rgb}{0.5,0.5,0.5}
\definecolor{codepurple}{rgb}{0.58,0,0.82}
\definecolor{backcolour}{rgb}{0.95,0.95,0.92}

\usepackage{listings}
\lstdefinestyle{mystyle}{
    backgroundcolor=\color{backcolour},   
    commentstyle=\color{codegreen},
    keywordstyle=\color{magenta},
    numberstyle=\tiny\color{codegray},
    stringstyle=\color{codepurple},
    basicstyle=\ttfamily\footnotesize,
    breakatwhitespace=false,         
    breaklines=true,                 
    captionpos=b,                    
    keepspaces=true,                 
    numbers=left,                    
    numbersep=5pt,                  
    showspaces=false,                
    showstringspaces=false,
    showtabs=false,                  
    tabsize=2
}
\usepackage[numbers,sort&compress]{natbib}
\usepackage{cleveref} 
\AtBeginEnvironment{appendices}{\crefalias{section}{appendix}}

\newcommand\brackets[1]
{\mathopen{}\left[#1\right]\mathclose{}}
\newcommand\parens[1]{\mathopen{}\left(#1\right)\mathclose{}}
\newcommand\braces[1]{\mathopen{}\left\{#1\right\}\mathclose{}}

\newcommand{\cmt}[1]{} 

\DeclareMathOperator*{\E}{\mathbb{E}}
\newcommand{\losszeta}{\mathcal{L}(\boldsymbol{\zeta})} 

\newcommand{\obj}{l_o}
\newcommand{\Obj}{L_o}

\newcommand{\psib}{\boldsymbol{\psi}}
\newcommand{\alphab}{\boldsymbol{\alpha}}

\newcommand{\phib}{\boldsymbol{\phi}}

\newcommand{\thetab}{\boldsymbol{\theta}}
\newcommand{\zetab}{\boldsymbol{\zeta}}

\newcommand{\bb}{\boldsymbol{b}}

\newcommand{\fb}{\boldsymbol{f}}

\newcommand{\mb}{\boldsymbol{m}}

\newcommand{\niu}{\mathrm{u}}
\newcommand{\niub}{\boldsymbol{\mathrm{u}}}

\newcommand{\xb}{\boldsymbol{x}}

\newcommand{\nixb}{\boldsymbol{\mathrm{x}}} 
\newcommand{\niXb}{\boldsymbol{\mathrm{X}}} 

\newcommand{\niz}{\mathrm{z}} 
\newcommand{\nizb}{\boldsymbol{\mathrm{z}}}

\newcommand{\smto}{\texttt{SMTO}}
\newcommand{\lcsmto}{\texttt{LC-SMTO}}

\usepackage{titling}
\thanksheadextra{}{}
\thanksheadextra{}{} 
\usepackage{lipsum} 

\title{Localized Physics-informed Gaussian Processes with Curriculum Training for Topology Optimization}
\date{\vspace{-5ex}}
\usepackage{authblk}
\author[1]{Amin Yousefpour}
\author[1]{Shirin Hosseinmardi}
\author[1]{Xiangyu Sun}
\author[1,2] {Ramin Bostanabad \thanks{Corresponding Author: Raminb@uci.edu}}
\affil[1]{Department of Mechanical and Aerospace Engineering, University of California, Irvine}
\affil[2]{Department of Civil and Environmental Engineering, University of California, Irvine}

\begin{document}
    \pagenumbering{arabic}
    \sloppy
    \maketitle
    \noindent \textbf{Abstract}\\
We introduce a simultaneous and meshfree topology optimization (TO) framework based on physics-informed Gaussian processes (GPs). Our framework endows all design and state variables via GP priors which have a shared, multi-output mean function that is parametrized via a customized deep neural network (DNN). The parameters of this mean function are estimated by minimizing a multi-component loss function that depends on the performance metric, design constraints, and the residuals on the state equations. Our TO approach yields well-defined material interfaces and has a built-in continuation nature that promotes global optimality. Other unique features of our approach include (1) its customized DNN which, unlike fully connected feed-forward DNNs, has a localized learning capacity that enables capturing intricate topologies and reducing residuals in high gradient fields, (2) its loss function that leverages localized weights to promote solution accuracy around interfaces, and (3) its use of curriculum training to avoid local optimality.  
To demonstrate the power of our framework, we validate it against commercial TO package COMSOL on three problems involving dissipated power minimization in Stokes flow.

    \section{Introduction} \label{sec intro}
TO offers a systematic approach to answer a fundamental engineering question: how should material be distributed within a given design domain to achieve optimal structural performance? Originally developed for mechanical design, TO has since expanded into various physical disciplines, including fluid dynamics, acoustics, electromagnetics, and optics, as well as their interdisciplinary applications.

Most conventional TO methods adopt a \textit{nested} framework. In these approaches, each design iteration is divided into two interlinked phases: an analysis phase and an update phase. During the analysis phase, the candidate design is evaluated by solving the governing equations (which are typically partial differential equations or PDEs) using numerical methods such as the finite element method (FEM) \cite{RN2050,fin2019structural}. The obtained responses are then used to update the structure for the next design iteration. The analysis-update iterations are continued until a convergence metric is met \cite{ben2024robust,stragiotti2024efficient}. While the nested strategy has been extensively used, its reliance on discretization introduces challenges such as mesh sensitivity and the need for spatial filtering to ensure mesh-independency. The discretization of the design domain necessitates some compromises, e.g., treating voids as regions with a small but nonzero stiffness. These compromises can limit the fidelity of the optimized designs, especially in problems involving complex multi-physics interactions or large deformations \cite{RN1986}. Some studies suggest meshfree methods as an alternative to address these issues \cite{RN2046,RN2048}, yet they often bring additional computational overhead and require ad hoc procedures to ensure stability. Also, nested TO methods can be computationally expensive due to repeated PDE solving. 

\textit{Simultaneous} TO methods have been introduced as an alternative to nested methods, aiming to integrate the analysis and design update steps into a unified optimization process \cite{haftka1985simultaneous,bendsoe1991new}. By embedding the governing equations directly within the optimization loop, these methods can potentially accelerate convergence. However, this direct coupling of physics and design variables increases computational complexity and memory demands, often limiting scalability. Moreover, most existing simultaneous approaches, like their nested counterparts, remain dependent on meshing techniques and inherit the associated challenges, including discretization errors and the need for density filtering to ensure numerical stability and mesh-independency. 

Some of these limitations are recently addressed in \cite{yousefpour2025simultaneous} where the authors propose a simultaneous and meshfree TO approach that combines the design and analysis steps into a single optimization loop. Their framework is formulated based on physics-informed GPs whose mean functions are parameterized via deep neural networks (DNNs). Specifically, GP priors are placed on all design and state variables to represent them via parameterized continuous functions. These GPs share a single DNN as their mean function but have as many independent kernels as there are state and design variables. Henceforth, we refer to this method as \smto.

While \smto~outperforms competing DNN-based TO approaches, it has three main limitations. First, it lacks a localized learning mechanism which makes it difficult to accurately solve the associated PDEs close to interfaces or in regions with high-gradient solutions. 
Second, \smto~ relies on estimating a density function which results in some gray area. While \smto~achieves less gray areas compared to traditional density-based methods, the generated density function is ultimately passed through a projection step which fails to enforce a completely binary material distribution. 
Third, the simultaneous nature of the approach may insufficiently prioritize the underlying governing equations while minimizing the loss function which depends on the objective, PDE residuals, and design constraints. This behavior can decelerate convergence or produce sub-optimal topologies. 

To address these limitations, we introduce a new TO approach with a localized learning capacity that also leverages a curriculum-based training strategy. These advancements, along with incorporating a level set function rather than a material density field,  enable our approach to reduce gray areas, converge faster, and solve more complicated TO problems. We denote our approach by \lcsmto~and introduce it in Section \ref{sec method} and then evaluate its performance against COMSOL on multiple benchmark problems in Section \ref{sec results}. We conclude the paper in Section \ref{sec conclusion} by summarizing our contributions and providing future research directions.

    \section{Proposed Methodology} \label{sec method}
TO aims to find the spatial material distribution that minimizes the scalar objective function $\Obj(\cdot)$ subject to a set of scalar constraints $C_i(\cdot) \leq 0, i= 1, ..., n_c$.  The material distribution is denoted by the binary spatial variable $\rho(\nixb)$ where $\nixb = \brackets{x, y, z}$ are the spatial coordinates. The governing physics of the problem are defined by state equations $R_i(\niub(\nixb))$ based on $n_u$ state variables $\niub(\nixb)= \brackets{\niu_1(\cdot), \cdots, \niu_{n_u}(\cdot)}$. In many practical cases, such as the examples presented in Section \ref{sec results}, $R_i(\niub(\nixb))$ are PDEs.

\lcsmto~embeds the state equations $R_i(\niub(\nixb))$ directly into the optimization formulation by imposing them as constraints. After this embedding, and by introducing an intermediate continuous level set function $\psi(\nixb)$, we obtain:
\begin{subequations}
    \begin{align}
        &\widehat\rho(\nixb), \widehat\niub(\nixb) = \underset{\psi\parens{\nixb},  \niub(\nixb)}{\argmin} \hspace{2mm}
        \Obj\parens{\niub\parens{\nixb}, \rho(\nixb)} = 
        \underset{\psi\parens{\nixb}, \niub(\nixb)}{\argmin} \hspace{2mm}
        \int_\Omega \obj\parens{\niub\parens{\nixb}, \rho(\nixb), \nixb} d\omega, 
        \label{eq sim_to_generic_obj}\\
        & \text{subject to:} \notag \\ 
        &  \qquad  C_1\parens{\rho(\nixb)} = \int_\Omega \rho(\nixb) d\omega - V = 0, 
        \label{eq sim_to_generic_c1} \\
        &  \qquad  C_i\parens{\niub\parens{\nixb}, \rho(\nixb)} = 
        \int_\Omega c_i\parens{\niub\parens{\nixb}, \rho(\nixb), \nixb} d\omega \leq 0, \hspace{2mm} i= 2, ..., n_c, 
        \label{eq sim_to_generic_c2} \\
        &  \qquad  R_i\parens{\niub\parens{\nixb}} = 
        \int_\Omega r_i\parens{\niub\parens{\nixb}, \nixb} d\omega = 0, \hspace{2mm} i= 1, ..., n_r,
        \label{eq sim_to_generic_c3} \\        
 &  \qquad \rho\parens{\nixb} = h\bigl(\psi(\nixb)\bigr)= 
    \begin{cases}
        0, & \psi(\nixb) < 0 \\
        1, & \psi(\nixb) \geq 0 
    \end{cases},\quad \forall \nixb \in \Omega.
    \label{eq sim_to_generic_c4}
    \end{align}
    \label{eq sim_to_generic}
\end{subequations}
where $\Omega$ denotes the design domain, $V$ is the desired volume fraction, and constraint $R_i\bigl(\niub(\nixb)\bigr)$ corresponds to the $i^\text{th}$ state equation or to the associated initial/boundary conditions (ICs and BCs). We assume that the objective function and all constraints can be represented as the integral of appropriately defined local functions which do \textit{not} have a nested structure, e.g., we have $C_i\bigl(\niub(\nixb), \rho(\nixb)\bigr)$ instead of $C_i\bigl(\niub(\rho(\nixb), \nixb)\bigr)$.

$\rho(\nixb)$ in Equation \ref{eq sim_to_generic} is defined by applying a Heaviside step function $h(\cdot)$ to scalar level set function $\psi(\nixb)$. The region where $\psi(\nixb) \ge 0$ is interpreted as \textit{void} and where $\psi(\nixb) < 0$ as \textit{solid}. Our choice is motivated by the level set approach \cite{RN2024,RN2037,RN2038} which implicitly represents boundaries via $\psi(\nixb)=0$. In our formulation, however, $\psi(\nixb)$ is determined directly and simultaneously with $\niub(\nixb)$. 

 The primary challenge of the simultaneous approaches lies in the complexity of solving the constrained optimization problem in Equation \ref{eq sim_to_generic}. This difficulty stems from the substantial differences in both the degree of nonlinearity and the scale between the objective function and the constraints. Similar to \smto~\cite{yousefpour2025simultaneous}, we address this challenge by parameterizing all dependent variables—specifically, $\niub(\nixb)$ and $\rho(\nixb)$—using a differentiable, multi-output function $\nizb(\nixb; \zetab)$ with parameters $\zetab$:
\begin{equation}
    \nizb(\nixb; \zetab) \coloneqq \brackets{\niu_1(\nixb), ..., \niu_{n_{\niu}}(\nixb), \rho\parens{\nixb}}.
    \label{eq z definition}
\end{equation}
For notational simplicity, we introduce the shorthands $\niz_{-1}(\nixb; \zetab) :=  h\bigl(\psi(\nixb)\bigr)=\rho\bigl(\nixb\bigr) \quad \text{and} \quad \nizb_{\sim1}(\nixb; \zetab) := \niub\bigl(\nixb\bigr)$. Having introduced this notation,  we now convert Equation \ref{eq sim_to_generic} into the following unconstrained form via the penalty method:
\begin{equation}
    \begin{aligned}
        \widehat\zetab = \underset{\zetab}{\argmin} \hspace{2mm} \losszeta
        &= \underset{\zetab}{\argmin} \hspace{2mm} 
        \Obj\parens{\nizb(\nixb; \zetab)} + 
        \mu_p \left( 
        \sum_{i=1}^{n_r}\alpha_i R_i^2\parens{\nizb_{\sim1}(\nixb; \zetab)}\right. \\
        & \left. + \alpha_{n_r + 1}C_1^2\parens{\niz_{-1}(\nixb; \zetab)} + \sum_{i=2}^{n_c}\alpha_{n_r+i} \max\parens{0, C_i^2\parens{\nizb(\nixb; \zetab)}} \right).
    \end{aligned}
    \label{eq penalty_opt_generic}
\end{equation}
where $\mu_p$ denotes the penalty factor and $\alphab = \brackets{\alpha_1, \cdots, \alpha_{n_r+n_c}}$ are weights. We refer the reader to \cite{yousefpour2025simultaneous} for details on adaptively updating the penalty factor and weights during optimization to ensure balanced contributions of each term to $\losszeta$. 
Once the minimization problem in Equation \ref{eq penalty_opt_generic} is solved, the designed topology can be obtained via $\niz_{-1}(\nixb; \widehat{\zetab})$.

To evaluate each individual term on the right-hand side of Equation~\ref{eq penalty_opt_generic} and then approximate it by summation:
\begin{subequations}
    \begin{align}
            & \Obj\parens{ \nizb(\nixb; \zetab)} = \int_{\Omega} \obj\parens{ \nizb(\nixb; \zetab), \nixb} d\omega \approx \sum_{i=1}^{n_{CP}} \obj\parens{\nizb(\nixb_i; \zetab) \omega_i},
        \label{eq sim_to_dis_lo}\\
           & R_i\parens{\nizb_{\sim1}(\nixb_i; \zetab)} = 
            \int_{\Omega} r_i\parens{\nizb_{\sim1}(\nixb; \zetab), \nixb} d\omega 
            \approx \sum_{j=1}^{n_{CP}} w(\nixb_j) r_i\parens{\nizb_{\sim1}(\nixb_j; \zetab)} \omega_j 
            = 0, \quad i= 1, \dots, n_r.
        \label{eq sim_to_generic_ri}\\
            & C_1\parens{ \niz_{-1}(\nixb; \zetab)} = 
            \int_{\Omega} c_1\parens{ \niz_{-1}(\nixb; \zetab), \nixb} d\omega - V 
            \approx \sum_{j=1}^{n_{CP}} \niz_{-1}(\nixb_j; \zetab) \omega_j - V 
            = 0.
        \label{eq sim_to_dis_c1}\\
           & C_i\parens{\nizb(\nixb; \zetab)} = 
            \int_{\Omega} c_i\parens{\nizb(\nixb; \zetab), \nixb} d\omega
            \approx \sum_{j=1}^{n_{CP}} c_i\parens{\nizb(\nixb_j; \zetab)} \omega_j
            \leq 0, \quad i= 2, \dots, n_c.
        \label{eq sim_to_generic_ci}
        \end{align}
\end{subequations}
where $\nixb_j$ denotes a set of $n_{CP}$ collocation points (CPs) distributed over the design domain and $w(\nixb_j)$ denotes a localized weighting function detailed in Section~\ref{subsec:L_Weighting}. 

We now revisit the parameterization of our search space, i.e., $\nizb(\nixb; \zetab)$. Since state variables inherently depend on the topology, it is logical to parameterize both $\niub(\nixb)$ and $\rho(\nixb)$ jointly by a single function. This unified representation must possess sufficient expressivity to characterize a wide variety of topologies and state distributions; otherwise, problem-specific representations would become necessary. Additionally, efficient gradient computations of $\nizb(\nixb; \zetab)$ with respect to both $\zetab$ and $\nixb$ are crucial, as gradients with respect to $\nixb$ are typically required to enforce $R_i\parens{\nizb_{\sim1}(\nixb_i; \zetab)}$. Finally, sharp interfaces and discontinuities can pose challenges to global representations, necessitating localized or adaptive strategies during optimization.

Motivated by these considerations, we adopt a localized, physics-informed GP for $\nizb(\nixb; \zetab)$. This choice automatically satisfies ICs and BCs; simplifying Equation~\eqref{eq penalty_opt_generic} as we can now exclude the corresponding penalties. Further details on our parameterization and its training are provided below. 
\begin{figure*}[!hbt]
    \centering
    \includegraphics[width=\linewidth]{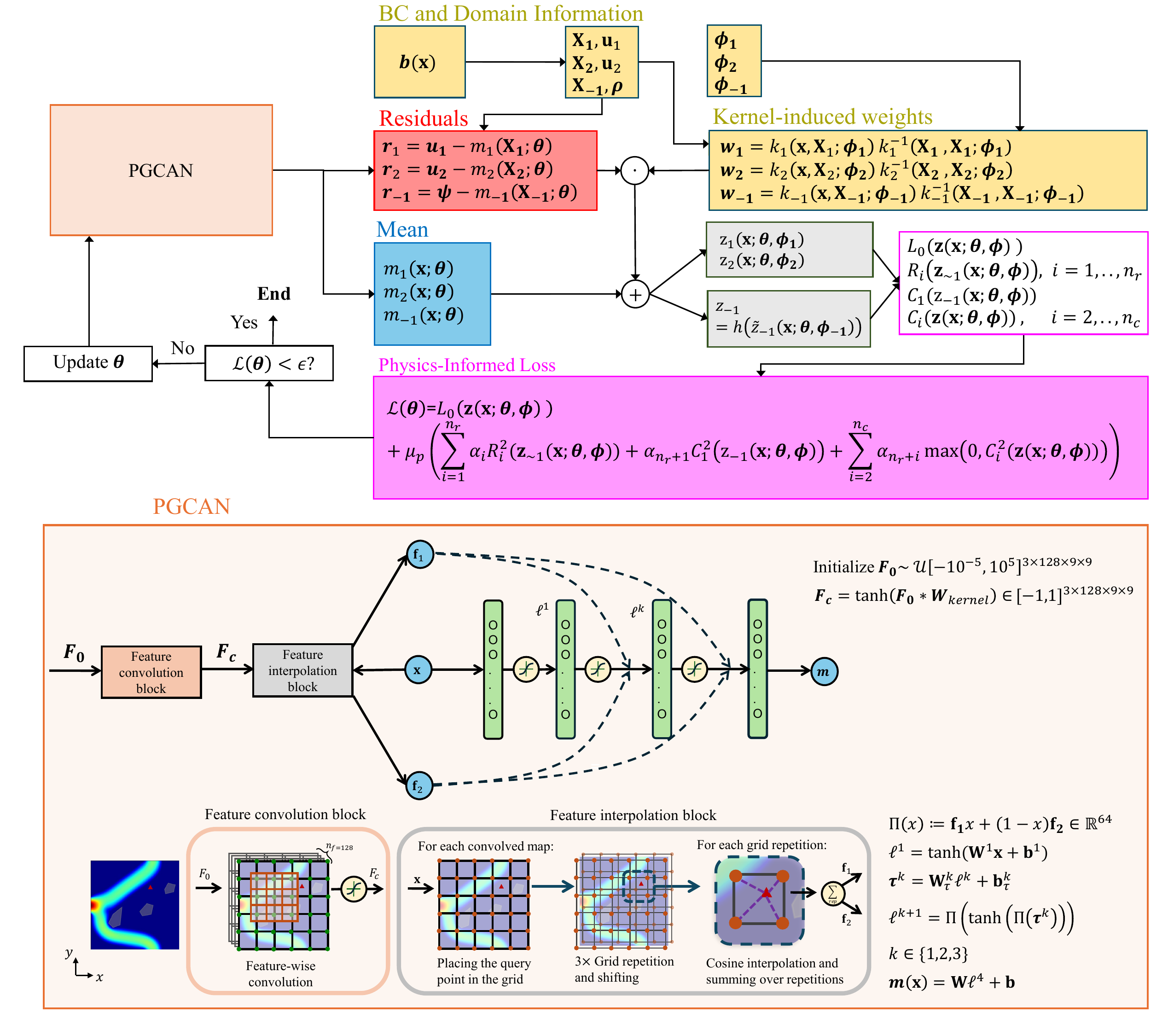}
    \caption{\textbf{ Simultaneous and meshfree topology optimization with localized features :} It is assumed that the structure has two state variables $\niub(\nixb) = \brackets{u_1(\nixb), u_2(\nixb)}$. The level set function $\psi(\nixb)$ indicates the material phase at any given point where negative/positive values correspond to solid/void. The covariance matrices ensure that the variables satisfy the boundary conditions while the parameters of the PGCAN (mean function) are optimized to minimize Equation \eqref{eq penalty_opt_generic}. In practice, we fix the length-scale parameters of all the kernels to $10^3$ and only optimize $\thetab$.}
    \label{fig flowchart}
\end{figure*}
\subsection{Parameterization of Design and State Variables} \label{subsec model description}
$\nizb(\nixb; \zetab)$ must satisfy the PDE system governing the problem. To understand how this requirement affects our parameterization, we consider the following generic boundary value problem:
\begin{subequations}
    \begin{align}
        &\mathcal{N}_{\nixb}[\niub(\nixb)] = \fb(\nixb), \quad \nixb \in \Omega,
        \label{eq generic pde}\\
        &\niub(\nixb) = \bb(\nixb), \quad \quad \nixb \in \partial \Omega,
        \label{eq generic bc}
    \end{align}
    \label{eq generic pde system}
\end{subequations}
where $ \partial\Omega $ denotes the boundary of $\Omega$, $\mathcal{N}_{\nixb}[\niub(\nixb)]$ are a set of differential operators acting on $\niub(\nixb)$, $ \fb(\nixb) = \brackets{f_1(\nixb), \cdots, f_{n_u}(\nixb)} $ is a known vector of functions, and the prescribed BCs on the state variables are characterized via $ \bb(\nixb) = \brackets{b_1(\xb), \cdots, b_{n_u}(\xb)} $. 

Equations~\eqref{eq generic pde} and~\eqref{eq generic bc} can both be incorporated into Equation~\eqref{eq penalty_opt_generic} as equality constraints but this causes two main challenges. First, the PDE and BC constraints differ significantly in scale, which requires careful weight adjustment in $\alphab$. Second, Equation~\eqref{eq penalty_opt_generic} has a continuation nature since the PDEs are not strictly satisfied early in the optimization. This continuation, however, should not include the BCs as the response of the structure and the PDE solution greatly depend on it. That is, the BCs must be strictly satisfied during the optimization iterations. 

To address these issues, we employ physics-informed GPs \cite{mora2024neural} which inherently satisfy BCs/ICs imposed on $\niub(\nixb)$. Specifically, we first put a GP prior on each of the $n_u$ state variables. As justified below, we endow these GPs with separate kernels $k_i(\nixb, \nixb'; \phib_i)$ but a single shared multi-output mean function. We parameterize this mean function by a customized deep neural network (DNN) that we call PGCAN (detailed in Section \ref{subsec:PGCAN}) and denote it via $\mb(\nixb; \thetab) = \brackets{m_1(\nixb; \thetab), \cdots, m_{n_u}(\nixb; \thetab)}$. $\thetab$ and $\phib_i$ are the parameters of the mean function and the $i^{th}$ kernel. 

Having defined the mean function and kernels, we now condition the $i^{th}$ GP on the data that is sampled from the BCs/ICs imposed on the $i^{th}$ state variable. The $i^{th}$ conditional distribution is again a GP \cite{RN332,RN1559} whose conditional mean at an arbitrary query point $\nixb^*$ in the domain is given by:
\begin{equation}
    \niz_i\parens{\nixb^*; \thetab, \phib_i} \coloneqq 
    \E\brackets{\niu_i^*|\niub_i, \nixb} = 
    m_i\parens{\nixb^*; \thetab} + 
    k_i\parens{\nixb^*, \niXb_i; \phib_i}k_i^{-1}\parens{\niXb_i, \niXb_i; \phib_i}
    \parens{\niub_i- m_i\parens{\niXb_i; \thetab}}
    \label{eq gp_conditional_mean}
\end{equation}

where $i=1, \dots, n_{\niu}$; $\niu_i^* = \niu_i(\nixb^*)$; $\niXb_i = \braces{\nixb^{(1)}, \dots, \nixb^{(n)}}$ and the corresponding outputs $\niub_i = \braces{\niu_i(\nixb^{(1)}), \dots, \niu_i(\nixb^{(n)})}$ represent data sampled from the BCs/ICs imposed on the $i^{th}$ state variable. 

The term $\niub_i - m_i(\niXb_i; \thetab)$ in Equation~\eqref{eq gp_conditional_mean} quantifies how accurately the mean function reproduces the imposed boundary data. This quantity is scaled by $k_i(\nixb^*, \niXb_i; \phib_i)k_i^{-1}(\niXb_i, \niXb_i; \phib_i)$ which ensures that the conditional mean $\niz_i(\nixb^*; \thetab, \phib_i)$ exactly interpolates the boundary data $\niub_i$, regardless of the chosen forms and parameters of $m_i(\nixb; \thetab)$ and $k_i(\nixb, \nixb'; \phib_i)$. This interpolation property allows us to exclude constraints associated with BCs/ICs from Equation~\eqref{eq penalty_opt_generic} as long as we sufficiently sample from BCs/IC. Thus, Equation~\eqref{eq gp_conditional_mean} serves as our proposed parameterization for the state variables.

The parameterization of the level set function closely follows the formulation provided in Equation~\eqref{eq gp_conditional_mean}, with two key modifications: $(1)$ boundary data are replaced by samples corresponding explicitly to the level set values, and $(2)$ the Heaviside step function $h(.)$ is applied to the parameterized level set function to obtain binary outputs. That is:
\begin{subequations}
    \begin{align}
        &\widetilde\niz_{-1}\parens{\nixb^*; \thetab, \phib_{-1}} \coloneqq 
        \E\brackets{\psi^*|\psib, \nixb} \notag \\
        &\quad= m_{-1}\parens{\nixb^*; \thetab} +  k_{-1}\parens{\nixb^*, \niXb_{-1}; \phib_{-1}}
        k_{-1}^{-1}\parens{\niXb_{-1}, \niXb_{-1}; \phib_{-1}}
        \parens{\psib- m_{-1}\parens{\niXb_{-1}; \thetab}} \\
        &\niz_{-1}\parens{\nixb^*; \thetab, \phib_{-1}}  = h\parens{\widetilde\niz_{-1}\parens{\nixb^*; \thetab, \phib_{-1}}},
    \end{align}
    \label{eq gp_conditional_mean_rho_12}
\end{subequations}
where $\widetilde{\niz}_{-1}(\nixb^*; \thetab, \phib_{-1})$\footnote{We employs the shorthand notation where the subscript ${n_{\niu}+1}$ is replaced by the subscript $-1$.} denotes the parameterized level set function, and ${\niz}_{-1}(\nixb^*; \thetab, \phib_{-1})$ represents its binarized form. $\niXb_{-1} = \braces{\nixb^{(1)}, \cdots, \nixb^{(n)}}$\footnote{For clarity and simplicity, we assume the density and state variables are known at $n$ specific locations; however, in practice, this number may differ across various variables.} denotes locations where the material phase is known and the corresponding level set is represented by $\psib = \braces{\psi(\nixb^{(1)}), \cdots,\psi(\nixb^{(n)})}$. By definition, the absolute magnitudes of the numerical values assigned to $\psib$ have no physical significance; rather, it is solely their signs that distinguish the different material phases. In this study, we assign $-0.5$ to points in the solid phase and $+0.5$ to those in the void phase. This symmetric assignment promotes numerical stability that may occur due to vanishing or exploding gradients. 

While many kernels are available \cite{yousefpour2024gp+}, we adopt a simplified Gaussian covariance function with fixed, large length-scale parameters to interpolate all boundary points. This choice narrows the optimization task exclusively to estimating the parameters of the mean function, i.e., $\thetab$. A detailed discussion on this kernel, including its mathematical formulation and the rationale behind its selection, is provided in \cite{yousefpour2025simultaneous}.

Figure~\ref{fig flowchart} schematically illustrates our proposed framework where a single multi-output mean function approximates both the displacement field and the level set function, that is $\niub(\nixb)$ and $\psi(\nixb)$, in the domain while independent kernels are used for each variable. Having a shared mean function with separate kernels offers three distinct advantages. First, it enables effective handling of heterogeneous data in scenarios where BCs for state variables and the level set function are available at different locations or substantially differ. Second, it significantly reduces computational overhead and memory usage as covariance matrices are constructed separately for each dependent variable. For example, given $m$ dependent variables each evaluated at $n$ boundary points, our method involves $m$ covariance matrices, each of size $n\times n$, rather than a single large covariance matrix of size $mn\times mn$.  Third, as detailed in Section \ref{subsec:PGCAN}, the shared mean function inherently encodes and preserves the correlation between state and design variables while having localized learning capacity.

\subsection{Mean Function: Localized Learning via PGCAN} \label{subsec:PGCAN}
Due to the importance of capturing localized topological features and minimizing PDE residuals around solid-fluid interfaces, we parameterize the mean function via Parametric Grid Convolutional Attention Network or PGCAN \cite{shishehbor2024parametric}. As shown in Figure \ref{fig flowchart}, PGCAN is a DNN with customized architecture that consists of an encoder and a decoder. 

The encoder parameterizes the design domain $\Omega$ via a structured grid that has trainable parameters on its vertices. This setup allows \lcsmto~to learn localized features because the parameters of a vertex are optimized based on their surrounding cells and not the entire design domain. The trainable grid is also convolved and slightly perturbed to effectively influence the query points even outside their immediate neighborhoods. This convolution followed by perturbation offers two primary benefits: (1) it introduces a natural multi-scale representation, akin to hierarchical feature extraction commonly employed in convolutional neural networks for computer vision tasks; (2) it relaxes the strict locality constraints inherent in traditional grid-based encodings, which prevents extreme gradients and mitigates overfitting. After these steps, the feature vectors $( \mathbf{f}_1, \mathbf{f}_2)$ corresponding to each query point are obtained via an $H^1$ interpolator and passed to the decoder for further processing.

The decoder is a relatively shallow neural network with $3$ hidden layers and $64$ neurons per layer. It leverages an attention mechanism to regulate the contributions of the encoder's features across multiple depths, which enhances gradient flow during training.

PGCAN is known to efficiently solve PDEs exhibiting locally high-frequency behaviors by adaptively prioritizing regions of higher complexity \cite{shishehbor2024parametric}, thereby addressing challenges associated with sharp material interfaces that commonly arise in TO.

\subsection{Localized PDE Residual Weighting} \label{subsec:L_Weighting}
While PGCAN helps with learning complex solution patterns (around solid-fluid interfaces as in the examples of Section \ref{sec results} or around stress concentration regions in compliance minimization problems), it is insufficient. This limitation of ML models in solving PDEs is well known in the literature \cite{jagtap2020extended}. 
 
To address this limitation, we assign different weights to the CPs in $\Omega$. Specifically, we leverage the fact that the design boundaries can be easily identified via $\psi(\nixb) \approx 0$ in each iteration. So, we increase the contribution of CPs which are close to the boundaries to the loss. To this end, we define the following distance-based weight:
\begin{equation}
w(\mathbf{x}) \;=\;
\begin{cases}
w_h - (w_h - w_l)\,\dfrac{d(\mathbf{x},\Gamma_0)}{\delta}, & d(\mathbf{x},\Gamma_0) < \delta,\\[6pt]
w_l, & d(\mathbf{x},\Gamma_0) \geq \delta.
\end{cases}
\label{eq local_weight}
\end{equation}
where $\Gamma_0$ denotes the zero-level set and $\delta$ is a threshold distance that identifies the points that are close to the interface. $w_l$ and $w_h$ represent the lower and upper weight values that are assigned to the points sufficiently far from the interface (where $d(\mathbf{x}, \Gamma_0) \geq \delta$) or exactly on the interface (where $d(\mathbf{x}, \Gamma_0) = 0$), respectively.

In Section~\ref{sec results}, we set $\delta = 0.1$, $w_l = 0.9$, and $w_h = 2$. With this choice, CPs satisfying $d(\mathbf{x}, \Gamma_0) < 0.1$ smoothly transition in weight from $0.9$ to $2$ as they approach the interface. Selecting $w_l$ less than $1$ ensures that the total magnitude of the weighted PDE residual remains comparable to the case without localized weighting. Consequently, scaling the PDE residual by $w(\mathbf{x})$ negligibly changes its scale. 

\subsection{Curriculum Training} \label{subsec CS}
Curriculum training is a strategy where a model is trained on progressively more and more difficult tasks \cite{bengio2009curriculum,soviany2022curriculum} and it has been successfully used in many applications 
\cite{jafarpour2021active,el2020student,shi2015recurrent,krishnapriyan2021characterizing,munzer2022curriculum}.

In \lcsmto, we leverage curriculum training to increase optimization stability and accuracy. Specifically, we apply it to the volume fraction constraint to enforce it gradually during the optimization. With this setup, our model initially learns the PDE solution over a single-phase material (i.e., $\rho(\nixb)=1$ everywhere). Then, we gradually apply the volume fraction constraint following a pre-defined schedule. 
In this work, we formulate the schedule as:
\begin{equation}
V_{\text{scheduled}}(i)
\;=\;
V_{\mathrm{target}}
\;+\;
\bigl(1- V_{\mathrm{target}}\bigr)
\biggl(1 - \frac{\min(b_n,\bigl\lfloor \tfrac{i}{b_s} \bigr\rfloor)}{b_n}\biggr)^{p_c},
\label{eq:TargetFracBlock}
\end{equation}
where $V_{\mathrm{target}}$ denotes the desired volume fraction, $i$ represents the optimization iteration, $b_s$ is the predefined block size, $p_c$ controls the rate of transition, and $\lfloor \cdot \rfloor$ denotes the floor function. Additionally, the total number of blocks is given by $b_n = \frac{i_{t}}{b_s}$ where $i_{t}$ is the iteration at which $V_{\mathrm{target}}$ equals $V_{\mathrm{target}}$ and remains constant thereafter.
For the studies in Section~\ref{sec results}, we use $b_s=20$, $i_{t}=4k$, and $p_c=1$ for all cases. Note that we choose $i_{t}$ significantly smaller than the total number of training iterations ($4k$ vs. $20k$) to ensure that the model reaches the desired volume fraction sufficiently early during training.

An alternative to our block-wise approach is to use a smooth polynomial-based schedule to gradually decrease the volume fraction in each iteration. However, our studies show that our schedule is more effective than continuous ones for two primary reasons (due to space limitations, we exclude these studies from the manuscript): $(1)$ in a block-wise approach, $V_{\text{scheduled}}(i)$ is constant within each block which gives the model sufficient time to fully learn the corresponding design dynamics without being continuously disrupted, and $(2)$ discrete adjustments at the end of each block reduce the risk of convergence to local minima. Regarding the second point we highlight that discrete transitions are frequently employed in optimization literature to avoid local minima. For instance, the use of step-decay learning rate schedules in stochastic optimization methods enhances convergence properties in highly non-convex problems  \cite{wang2021convergence}.

    \section{Results and Discussions} \label{sec results}


\begin{figure*}[!b]
    \centering
        \centering
        \includegraphics[width=1.00\columnwidth]{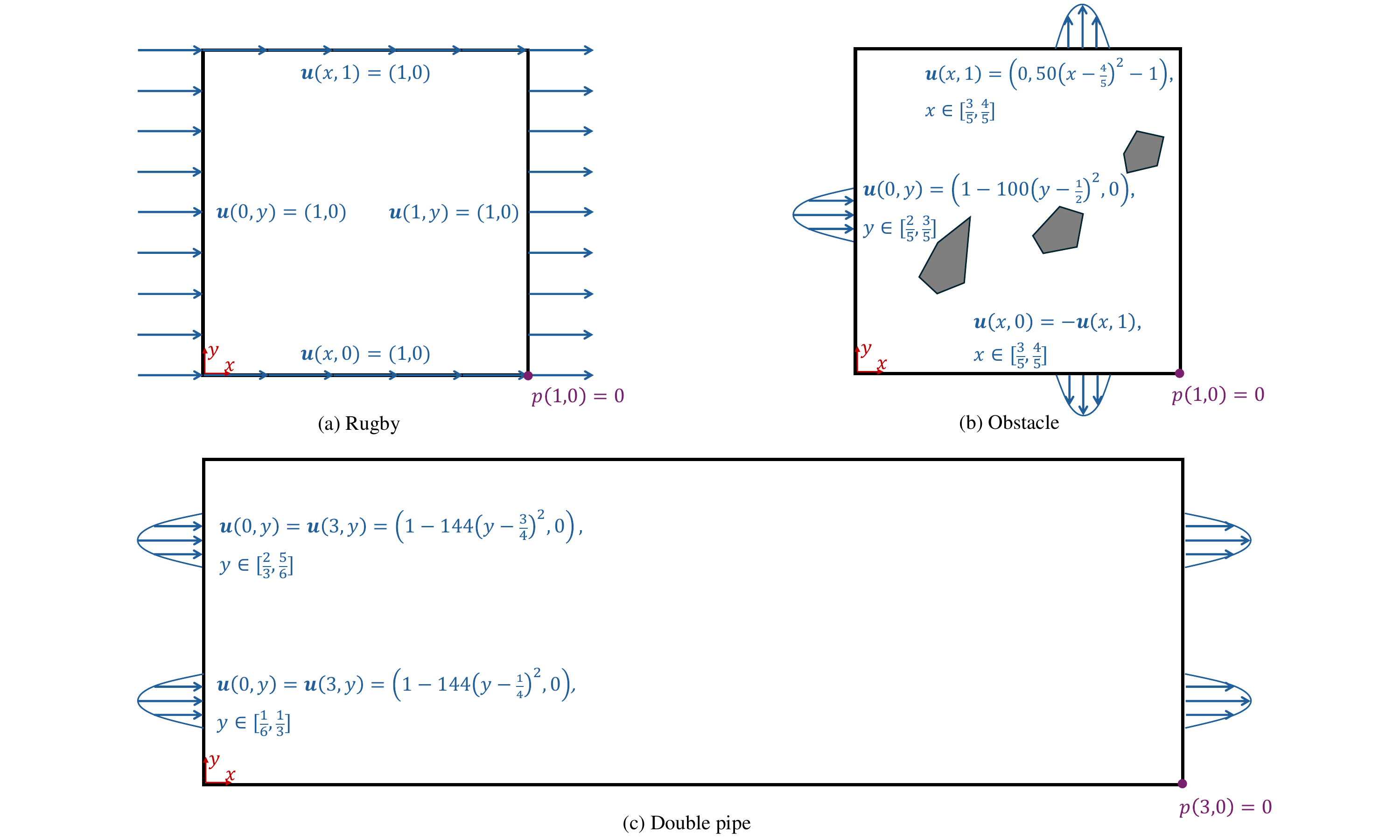}
        \label{fig:sub1}
    \caption{\textbf{Design domain and the imposed boundary conditions:} The Rugby and Obstacle domains are square, whereas the Double Pipe domain is a $3\times1$ rectangle.}
     \label{fig three problems}
\end{figure*}


We study 3 problems where the objective is to determine the spatial material distribution that minimizes the flow’s dissipated power under a predefined volume constraint. We consider the steady-state flow of an incompressible Newtonian fluid through porous media. The flow is governed by Brinkman equations \cite{brinkman1949calculation}: 
\begin{subequations}
    \begin{align}
        &  -\nabla p+\mu \nabla^2 \boldsymbol{u} - \mu \kappa^{-1} \boldsymbol{u} = 0
        \label{eq  momentum}\\
        & \nabla \cdot \boldsymbol{u} = 0
        \label{eq  continuity}.
    \end{align}
    \label{eq  Brinkman}
\end{subequations}
where $\mu$ denotes the dynamic viscosity, $\kappa$ denotes the permeability of the porous medium, $p$ is the pressure, and $\boldsymbol{u}$ is the velocity field. The model assumes negligible inertial effects due to low Reynolds number conditions and incorporates the resistive forces from the porous medium in accordance with Darcy's law. Gravitational effects are also neglected in the present analysis.
Assuming \(\mu = 1\) in a two-dimensional domain, the objective function is defined as:
\begin{equation}
    \mathcal{J}= \frac{1}{2} \int_{\Omega}\left( \nabla \boldsymbol{u}: \nabla \boldsymbol{u} +\kappa^{-1} \left \| \boldsymbol{u} \right \| ^2\right) dx dy ,
    \label{eq  objective}
\end{equation}
where the design variable $\rho$ is related to $\kappa$ by \cite{RN1986}:
\begin{equation}
    \kappa^{-1}(\rho)= {\kappa^{-1}}_{max} + \left({\kappa^{-1}}_{min} - {\kappa^{-1}}_{max}\right) \rho  \frac{1 + q}{\rho+q},
    \label{eq  alpha_ours}
\end{equation}
with ${\kappa^{-1}}_{max} = 2.5 \times 10^4 , \kappa^{-1}_{min} =  2.5 \times 10^{-4}$, and $q = 0.1$. The optimization problem is subject to fluid volume fraction: 
\begin{equation}
    \int_{\Omega}\rho dx dy = V_{\mathrm{target}}, 
    \label{eq  constraint}
\end{equation}
where $V_{\mathrm{target}}$ is equal to $0.9, 0.3$, and $\sfrac{1}{3}$ for our three test cases which are named as Rugby, Obstacle, and Double pipe, respectively. We demonstrate the BCs corresponding to each of these cases in Figure \ref{fig three problems} where domain boundaries (excluding inlets/outlets) represent no-slip walls.

For \lcsmto, we use the following settings to minimize Equation \eqref{eq objective} for each of the cases in Figure \ref{fig three problems}.
To ensure the BCs are accurately captured while the computational costs of matrix multiplications in Equations~\eqref{eq gp_conditional_mean},~\eqref{eq gp_conditional_mean_rho_12} 
are minimized, we sample $n_{bc} = 25$ points from the BCs on each side of the domain.
For calculating the PDE residuals, we sample the domain with $n_{cp} = 10k$ CPs for the Rugby and Obstacle examples and $n_{cp} = 30k$ for the Double pipe example. The CPs are positioned on a regular grid to facilitate the FD-based accelerations described in \cite{yousefpour2025simultaneous}. We train our models using the Adam optimizer with an initial learning rate of $0.001$ which is reduced by a factor of $0.75$ four times during $20k$ epochs.

In COMSOL, we set up the three problems via the Brinkman Equations (br) interface and the density model for the control variable field $\rho({\boldsymbol{\mathrm{x}}}) \in [0,1]$. 
The domain is meshed with Q2-Q1 elements (i.e., first order for pressure and $\rho$, and second order elements for the velocity field). Each element is a $0.01 \times 0.01$ square, similar to the CP setups in \lcsmto.
We use COMSOL's built-in Method of Moving Asymptotes (MMA), coupled with the Adjoint Gradient Method and Parallel Direct Sparse Solver Interface (PARDISO). We define the stop criteria as $0.05$ optimality tolerance and $100$ maximum iterations (whichever is first met). Additionally, the constraint penalty factor is set to $10^7$ for Rugby and $10^5$ for other problems.

\begin{figure*}[!t]
    \centering
        \centering
        \includegraphics[width=1.00\columnwidth]{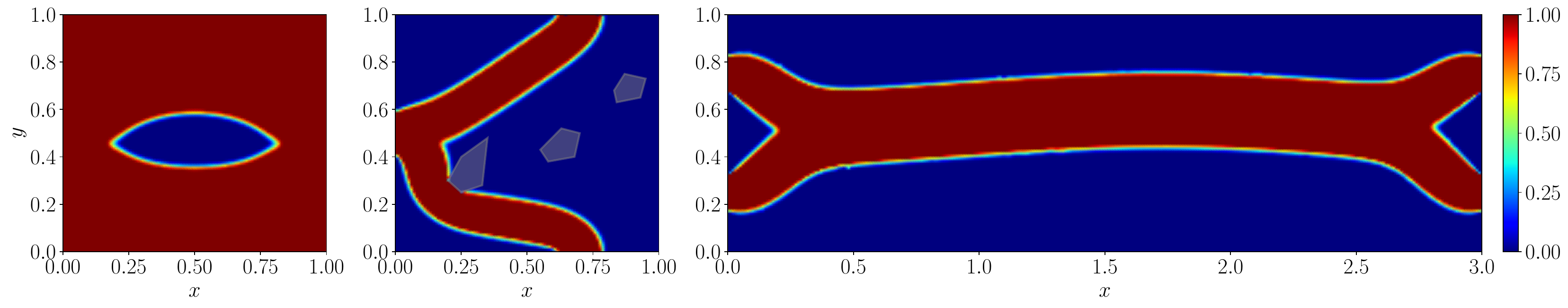}
        \label{fig:sub1}
    \caption{\textbf{COMSOL median optimal topologies:} The topologies corresponding to the median $\mathcal{J}$ out of $5$ random initializations are depicted for Rugby, Obstacle, and Double pipe problems.}
     \label{fig comsol top}
\end{figure*}

We initialize the density field in COMSOL randomly and find the optimum topology in five random repetitions. The median optimal topology for each problem is illustrated in Figure \ref{fig comsol top}. These results are discussed in detail in the following subsections and here we merely note that the optimum topologies in the Rugby and Double pipe examples are supposed to be symmetric, a property that is not exactly achieved by COMSOL.

\subsection{Results of Comparative Studies} \label{subsec:summary_results}

We evaluate the performance of \lcsmto~against \smto~and COMSOL. Table \ref{tab: results-summary} summarizes the median objective values and computational costs obtained from five independent repetitions for each method. Statistical details pertaining to the five repetitions are presented in Table \ref{tab stats}.

\begin{table}[!b]
    \caption{\textbf{Summary of results:} Median of final objective function value ($\mathcal{J}$) and computational cost (Time in sec) across $5$ repetitions.}
    \centering
    \renewcommand{\arraystretch}{1.1} 
    \small 
    \setlength\tabcolsep{1.5pt} 
    
    \begin{tabular}{l|c c|c c|c c}
    \hline
    \textbf{Problem} & \multicolumn{2}{c|}{\textbf{\lcsmto}} & \multicolumn{2}{c|}{\textbf{\smto}} & \multicolumn{2}{c}{\textbf{COMSOL}} \\ 
    & $\mathcal{J}$ & Time (sec)& $\mathcal{J}$ & Time (sec) & $\mathcal{J}$ & Time (sec) \\ \hline
    \textbf{Rugby} & 13.629 & 1332 & 14.176 & 1284 & 13.995 & 461 \\ \hline
    \textbf{Obstacle} & 6.509 & 1336 & 7.949 & 1245 & 7.094 & 246 \\ \hline
    \textbf{Double pipe} & 26.757 & 2558 & 24.600 & 2906 & 27.627 & 2225 \\ \hline
    \end{tabular}
    \label{tab: results-summary}
\end{table}

\begin{table}[!t]
    \caption{\textbf{Statistics of $\mathcal{J}$ associated with \lcsmto, \smto, and COMSOL with random initialization:} Our models' statistical metrics are gathered across $5$ training repetitions while COMSOL's stats are provided for  $5$ random initializations.}
    \centering
    \begin{tabular}{lccccc}
        \toprule
        \textbf{Problem} & \textbf{mean} & \textbf{median} & \textbf{std} & \textbf{min} & \textbf{max}\\
        \midrule
        \multicolumn{6}{c}{\textbf{\lcsmto~(proposed approach)}}\\        
        \midrule
        \textbf{Rugby} & 13.62 & 13.63 & 0.08 & 13.53 & 13.72 \\
        \textbf{Obstacle} & 6.39 & 6.51 & 0.31 &5.87 & 6.61 \\
        \textbf{Double pipe} & 26.39 & 26.76 & 0.93 & 24.76 & 27.08 \\
        \midrule
        \multicolumn{6}{c}{\textbf{\smto}}\\        
        \midrule
        \textbf{Rugby} & 14.11 & 14.18 & 0.45 & 13.58 & 14.76 \\
        \textbf{Obstacle} & 7.99 & 7.95 & 0.21 & 7.74 & 8.31 \\
        \textbf{Double pipe} & 26.17 & 24.60 & 5.31 & 22.76 & 35.53 \\
        \midrule
        \multicolumn{6}{c}{\textbf{COMSOL}}\\
        \midrule
        \textbf{Rugby} & 14.21 & 13.99 & 0.41 & 13.96 & 14.93 \\
        \textbf{Obstacle} & 7.10 & 7.09 & 0.01 & 7.09 & 7.11 \\
        \textbf{Double pipe} & 27.51 & 27.63 & 0.22 & 27.21 & 27.73 \\
        \bottomrule
    \end{tabular}
    \label{tab stats}
\end{table}


In Rugby, \lcsmto~consistently achieves the lowest objective function value while having the smallest standard deviation, underscoring its robustness. As shown in Section \ref{subsec: topology_evolution}, the designed topologies via \lcsmto~are also more symmetric compared to COMSOL.

\begin{figure*}[!b]
    \centering
        \centering
        \includegraphics[width=1.0\columnwidth]{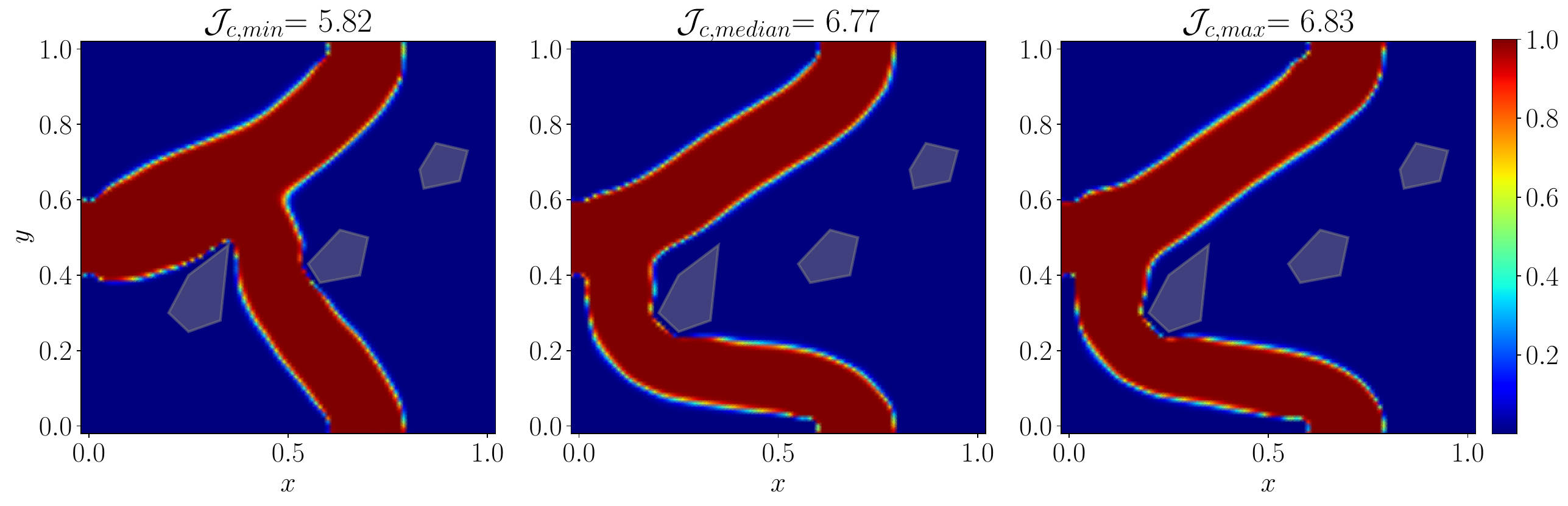}

    \caption{\textbf{Different topologies obtained for the Obstacle problem:} Designed topologies by \lcsmto~are imported in COMSOL to obtain $\mathcal{J}$ via FEA. Minimum and maximum values are relatively close but result in different topologies.}
     \label{fig obs}
\end{figure*}

In the Obstacle problem we observed that \lcsmto~ yields different topologies in $2$ out of $5$ training repetitions compared to COMSOL, despite close final $\mathcal{J}$ values, as reflected in the small standard deviation reported in Table \ref{tab stats}. This highlights that the standard deviation alone may not fully capture the diversity of solutions since topologies with distinct layouts can yield comparable objective values. To validate the performance of these alternative designs, we imported the final topologies into COMSOL, solved the flow and recomputed the dissipated power, denoted by $\mathcal{J}_c$. This post-analysis confirmed that the two alternative designs correspond to slightly more favorable solutions, with a small reduction in dissipated power compared to other instances.
Notably, COMSOL never generated such topologies in any of its optimization runs, suggesting that our approach more effectively explores the design space. The topologies corresponding to the minimum, median, and maximum $\mathcal{J}_c$ values are illustrated in Figure \ref{fig obs}.


For the Double pipe example, \lcsmto~achieves slightly lower median and mean $\mathcal{J}$ values compared to COMSOL. In addition, as shown in Section \ref{subsec: topology_evolution}, the topologies obtained via \lcsmto~are more symmetrical than those designed via COMSOL. Regarding \smto, we note that it occasionally converges to good minima but fails to consistently obtain a merged pipe in the middle of the design domain. This behavior results in higher standard deviation and maximum values of dissipated power for \smto~in Table \ref{tab stats}. 

While there are differences in terms of hardware and software, we find it useful to compare the computational costs of our approach against COMSOL. We observe in Table \ref{tab: results-summary} that COMSOL is the fastest method in the Rugby and Obstacle benchmarks. However, in the Double pipe problem, all models incur fairly similar computational costs. Moreover, runtimes for \lcsmto~and \smto~are comparable across all benchmarks which indicates that the contributions of this paper negligibility increase the computational costs. 

\subsection{Convergence and Topology Evolution} \label{subsec: topology_evolution}
\begin{figure}[!b]
    \centering
        \centering
        \includegraphics[width=1.00\columnwidth]{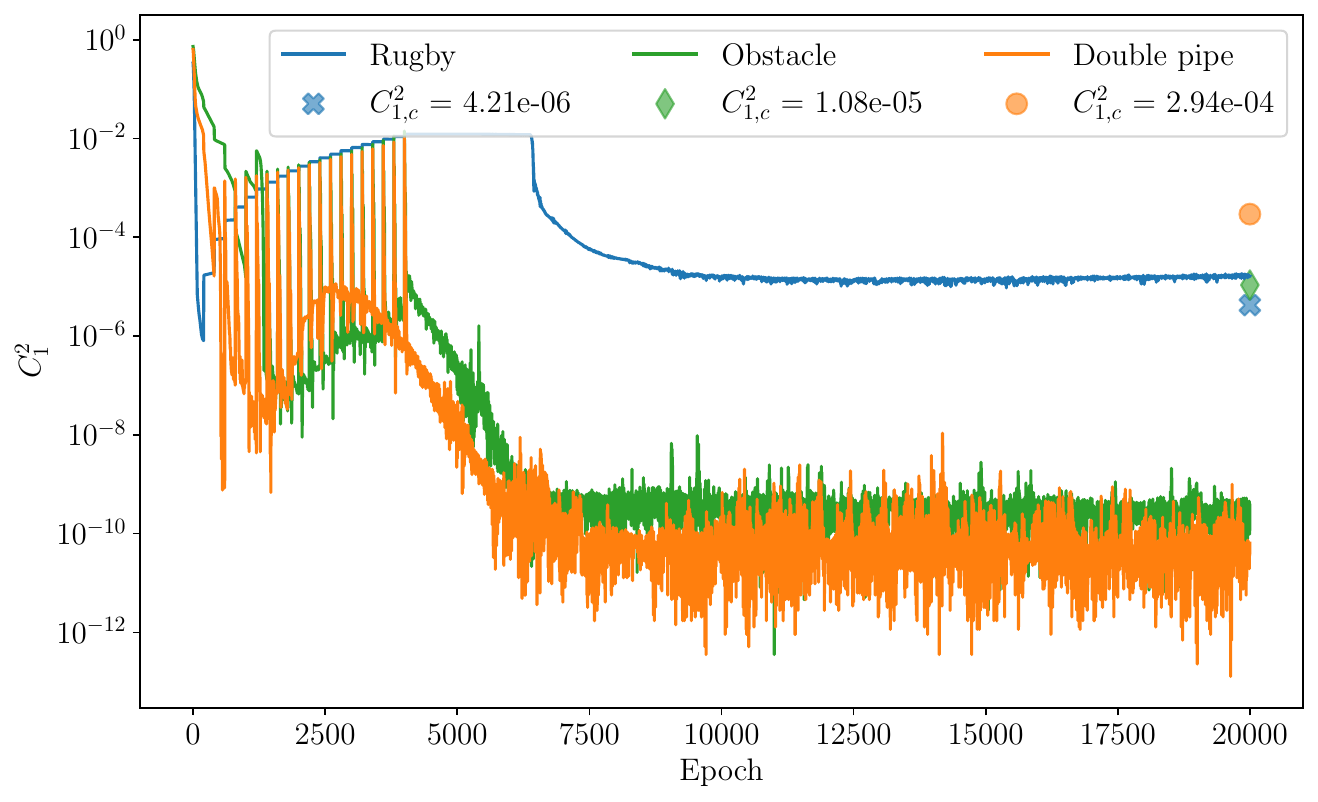}

    \caption{\textbf{Volume loss history:} Each curve represent the median across $5$ training repetitions. For comparison, we also provide the volume loss error obtained by COMSOL for its median optimal topology, see markers at $20k$.}
     \label{fig vol-loss}
\end{figure}

In this section, we analyze the loss histories and evolution of the optimum topologies during the optimization. 
Figure~\ref{fig vol-loss} shows the evolution of the volume loss error, i.e., $C_1^2$  in Equation \eqref{eq sim_to_generic_ri}, for the three problems where each curve represents the median loss over five independent training repetitions. For comparison, we also compute  $C_1^2$ for the median optimal topologies obtained by COMSOL (see the markers at $20k$ epoch). We notice that \lcsmto 's converged values are $4-5$ orders of magnitude smaller than those of COMSOL, except for the Rugby example where a stricter penalty constraint factor was applied in COMSOL.

We observe in Figure~\ref{fig vol-loss} that $C_1^2$ fluctuates substantially before $5k$ epochs. This trend is due to the incremental adjustment of the volume fraction constraint as described in Section~\ref{subsec CS}. The gradual increase in complexity introduces spikes in the volume constraint violation which is especially notable in the Double pipe example. These fluctuations, however, approximately die out after about $7k$ epochs. 


\begin{figure*}[!h]
    \centering
    \begin{subfigure}[b]{1\textwidth}
        \centering
        \includegraphics[width=1.0\columnwidth]{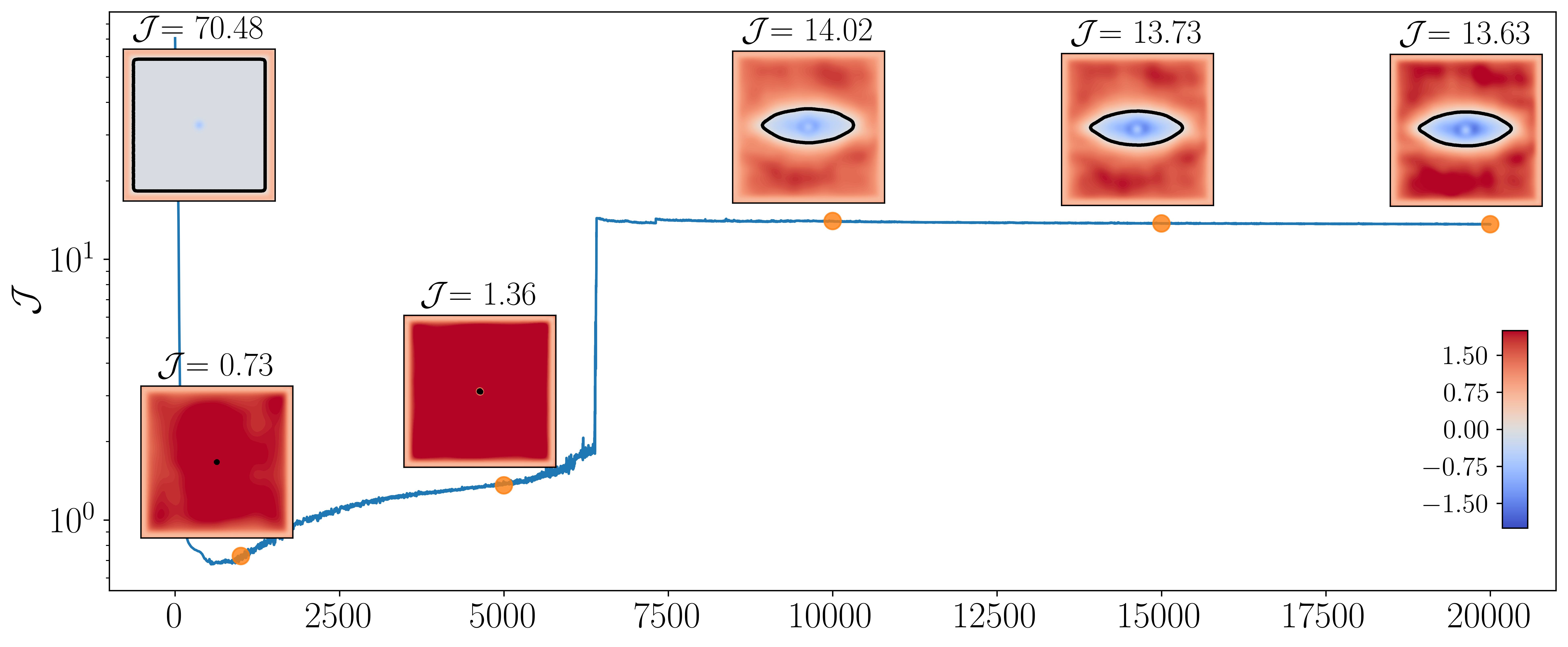}
    \end{subfigure}
    \begin{subfigure}[b]{1\textwidth}
        \centering
        \includegraphics[width=1.0\columnwidth]{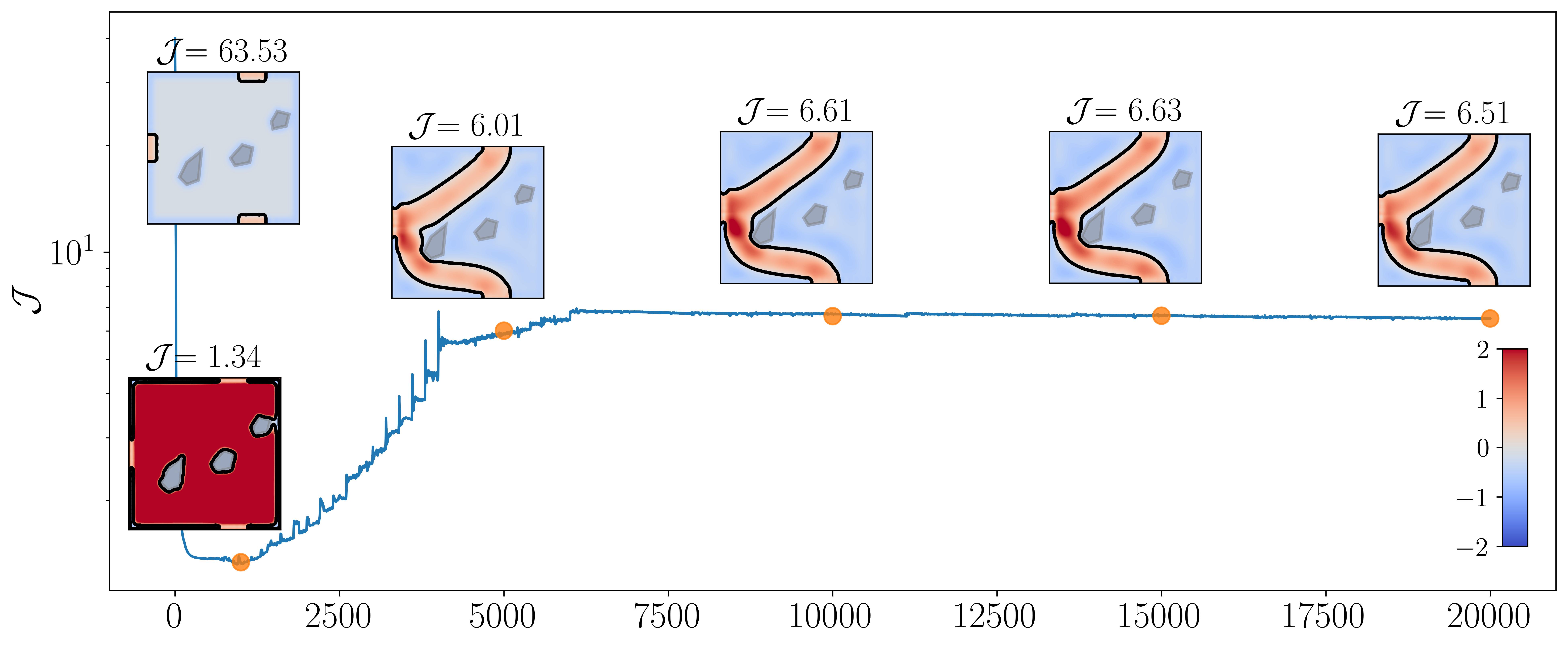}
    \end{subfigure}
    \begin{subfigure}[b]{1\textwidth}
        \centering
        \includegraphics[width=1.0\columnwidth]{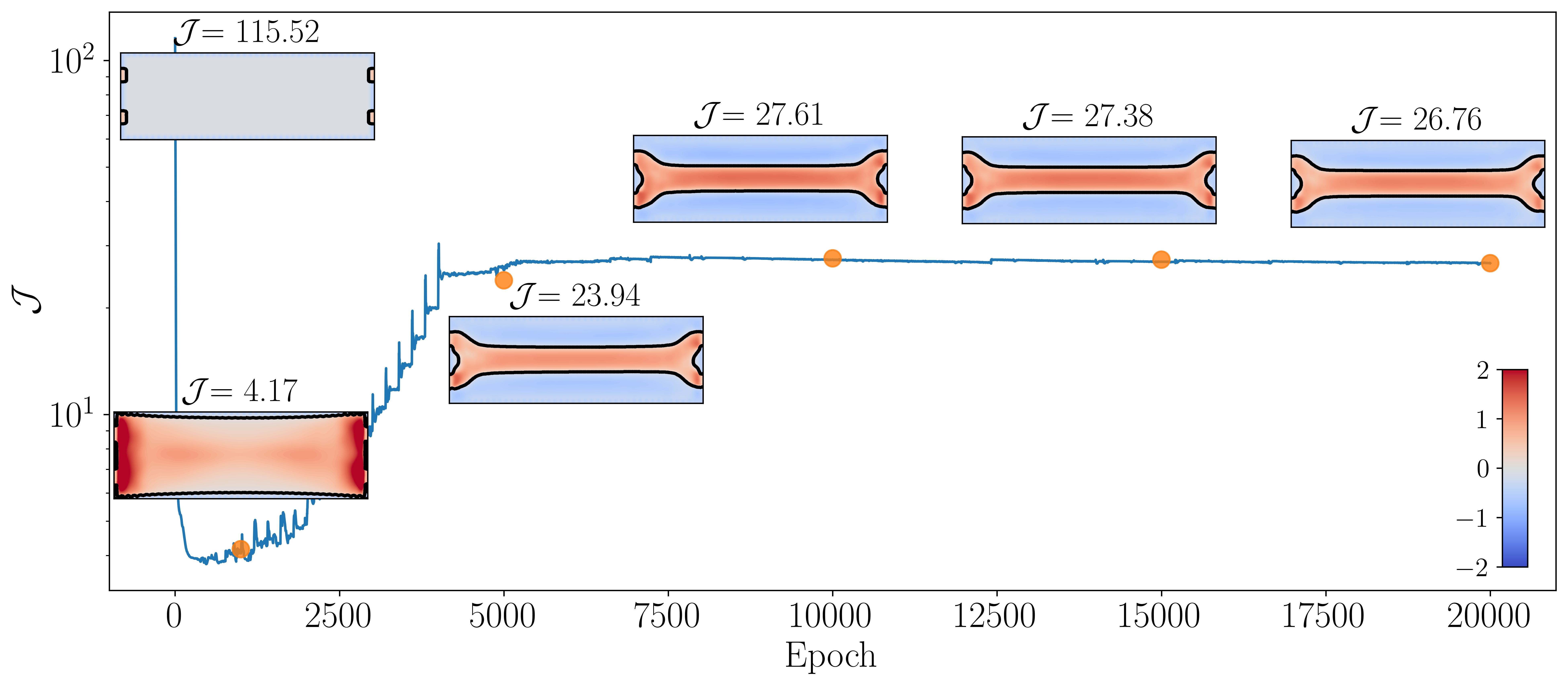}
    \end{subfigure}
    
    \caption{\textbf{Topology evolution across epochs:} The evolution of $\rho(\nixb)$ is visualized with respect to the objective function over $20k$ epochs. All quantities are the median across $5$ repetitions.}
    \label{fig rho evolution}
\end{figure*}

We next analyze the training dynamics of \lcsmto~by plotting the evolution of the objective function across $20k$ epochs in Figure \ref{fig rho evolution}. At epochs $1, 1k, 5k, 15k,$ and $20k$ we also visualize the median topologies from $5$ independent runs. In all examples, we observe that the optimum topologies are obtained well before $10k$ epochs after which neither the topologies nor the objective function values noticeably change. Interestingly, it takes \lcsmto~more iterations to converge in Rugby ($~7k$ epochs) than in the other two examples ($~4k$ epochs) even though the former seems to have a simpler optimal topology. 


In all cases in Figure \ref{fig rho evolution} $\mathcal{J}$ is high at first but then it rapidly decreases within the first few thousand epochs. These designs heavily violate the volume fraction constraint which is initially relaxed due to the curriculum training strategy and its adaptive penalty coefficient.
A comparison between the final topologies obtained by \lcsmto~(at iteration $20K$) and COMSOL’s median solutions in Figure~\ref{fig comsol top} demonstrates that \lcsmto~achieves superior symmetry (for example, COMSOL occasionally positions the Rugby slightly to the top or bottom of the domain).

\begin{figure*}[!b]
    \centering

    \includegraphics[width=1.00\columnwidth]{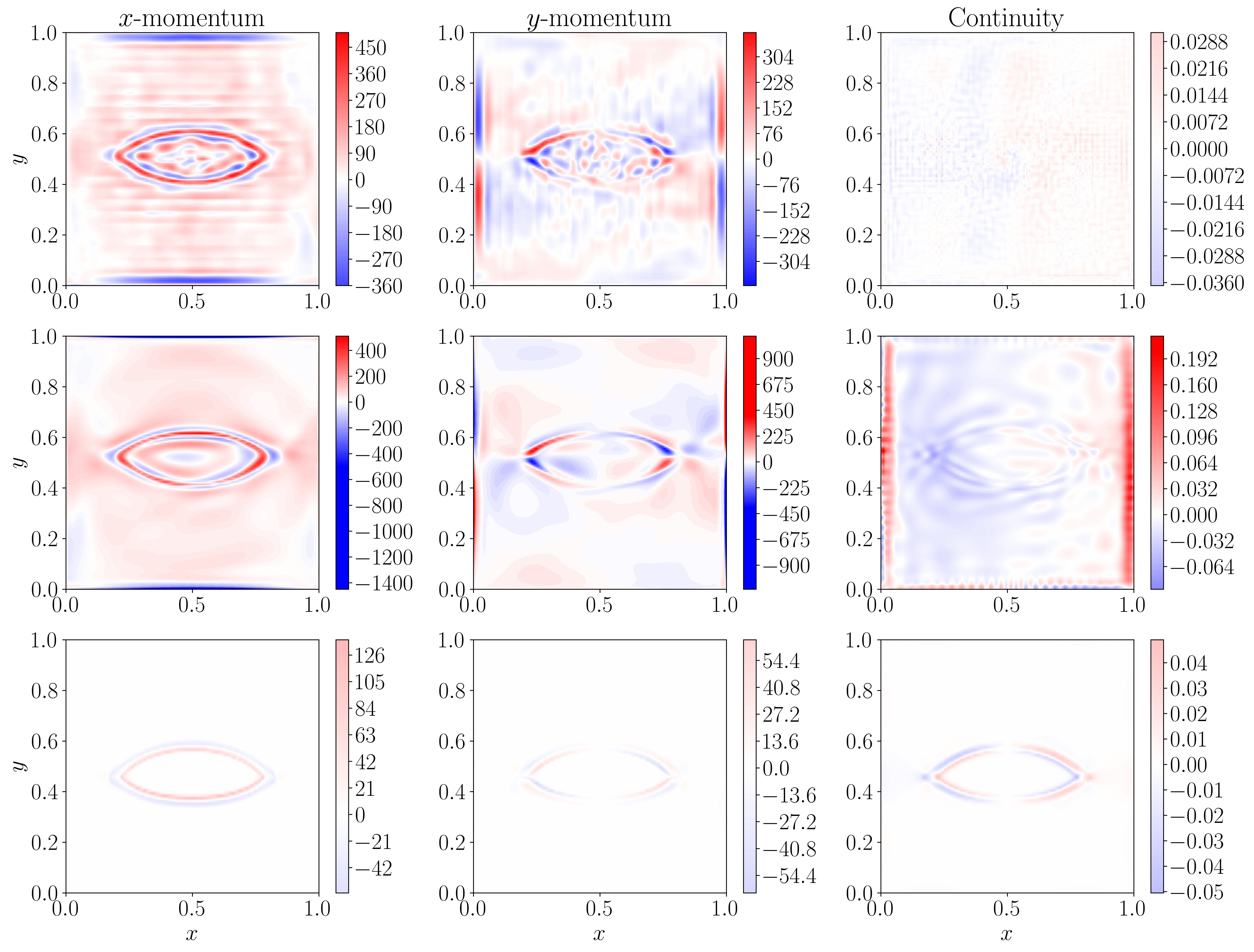}
        
    \caption{\textbf{Residual maps for Rugby's median optimal topology:} The top, middle, and bottom rows correspond to \lcsmto, \smto, and COMSOL.}
    \label{fig: res-map}
\end{figure*}
\subsection{Evaluation of PDE Residuals} \label{subsec residuals}

Even though solving the governing PDEs with strictly high accuracy might not be necessary while searching for the optimal topology, improved PDE solutions offer better search directions and lead to better designs as shown in the previous subsections where \lcsmto~consistently outperforms \smto. To compare the accuracy of the three methods in solving the PDE system in Equation \ref{eq  Brinkman}, we visualize the residual plots. 

Figure \ref{fig: res-map} displays Rugby's residual maps corresponding to the momentum and continuity equations for the median optimal topologies of \lcsmto~(top), \smto~(middle), and COMSOL (bottom). We observe that in the top row the maximum residual values near solid-fluid interfaces are approximately one order of magnitude smaller than those in the middle row. We attribute this improvement to the enhanced efficiency in solving the PDEs which is provided by $(1)$ PGCAN with its localized features, and $(2)$ the localized weighting mechanism which is applied after $9k$ training epochs. Regarding the second point, our localized weighting mechanism prioritizes critical points in later training stages and enables the model to capture their dynamics more accurately. In contrast, \smto~(middle row) allocates training effort more evenly, without targeted refinement in critical regions.

As expected, COMSOL (bottom row) achieves the smallest residuals overall, demonstrating superior accuracy in solving the weak form of the underlying PDE system via the FEM. Nonetheless, residual concentrations are still present near solid-fluid interfaces, suggesting that even high-fidelity FEM solutions experience some difficulty in fully resolving sharp transitions.

    \section {Conclusions and Future Directions} \label{sec conclusion}
We introduce a TO approach based on the framework of localized GPs that also leverages a curriculum-based training strategy. In our framework we solve the state equations while simultaneously estimating the level-set function that identifies the topology boundaries. Our approach has two features to capture local features regarding either the topology or the PDE solution characteristics. First, we parameterize the mean function of the GPs with PGCANs which encodes the design space via a set of features. Second, we develop a localized weighting mechanism to identify critical regions in the design space and, in turn, improve our model's accuracy in learning the underlying physics in these regions. 

In Section \ref{sec results} we consider $3$ problems where the goal is to identify the topology that minimizes the flow's dissipated power subject to a pre-defined volume constraint. We compare the resulting topologies obtained by our approach with those produced by \smto~and COMSOL. The results demonstrate three main advantages of our proposed method: 
$(1)$ it consistently outperforms \smto,
$(2)$ compared to COMSOL, it provides more optimal topologies in terms of the performance metric (dissipated power) or structural features (e.g., symmetry) and
$(3)$ it yields sharp interfaces.
 
In this work, we incorporate the strong form of the governing equations into the loss function. However, employing the weak form of the governing equations (as done in COMSOL) offers two notable advantages: $(1)$ it relaxes the smoothness requirements imposed on the solution, and $(2)$ it ensures a symmetric stiffness matrix in compliance minimization problems. We plan to investigate this promising extension in our future works.

\section*{Acknowledgments}
We appreciate the support from Office of the Naval Research (grant number $N000142312485$) and the National Science Foundation (grant numbers $2238038$ and $2525731$).

    \pagebreak
    \bibliography{01_Ref_r2}
\end{document}